\documentclass[10pt]{article} 

\usepackage[preprint]{rlj} 

\usepackage{amssymb}            
\usepackage{amsmath}
\usepackage{mathtools}          
\usepackage{mathrsfs}           
\usepackage{graphicx}           
\usepackage{subcaption}         
\usepackage[space]{grffile}     
\usepackage{url}                
\usepackage{lipsum}             

\usepackage[ruled,linesnumbered]{algorithm2e}

\SetAlFnt{\normalsize}
\SetAlCapFnt{\normalsize}
\SetAlCapNameFnt{\bfseries}
\SetAlgoNoLine
\DontPrintSemicolon
\SetKwFor{ForEach}{for each}{do}{end for}

\title{Gradient TD($\lambda$) for Non-Linear Function Approximation}

\setrunningtitle{Gradient TD($\lambda$) for Non-linear Function Approximation} 

\author{Sujeet Lahane}

\emails{sujeetlahaneqrs@gmail.com}

\affiliations{}

\contribution{
    Provide a succinct but precise list of the contribution(s) of the paper. Use contextual notes to avoid implications of contributions more significant than intended and to clarify and situate the contribution relative to prior work (see the examples below). If there is no additional context, enter ``None''. Try to keep each contribution to a single sentence, although multiple sentences are allowed when necessary. If using complete sentences, include punctuation. If using a single sentence fragment, you may omit the concluding period. A single contribution can be sufficient, and there is no limit on the number of contributions. Submissions will be judged mostly on the contributions claimed on their cover pages and the evidence provided to support them. Major contributions should not be claimed in the main text if they do not appear on the cover page. Overclaiming can lead to a submission being rejected, so it is important to have well-scoped contribution statements on the cover page.
    }
    {
    None
    }

\contribution{
    The submission template for submissions to RLJ/RLC 2025
    }
    {
    Built from previous RLC/RLJ, ICLR, and TMLR submission templates
    }

\contribution{
    \textit{{[}Example of one contribution and corresponding contextual note for the paper ``Policy gradient methods for reinforcement learning with function approximation'' \citep{Sutton2000}.]}\\ This paper presents an expression for the policy gradient when using function approximation to represent the action-value function.
    }
    {
    Prior work established expressions for the policy gradient without function approximation \citep{Williams1992}.
    }


\summary{

Maei et al introduced Gradient Temporal Difference Learning algorithms, with stronger theoretical backings, including convergence proofs in off-policy learning with non-linear function approximation. Further they extend the findings in form of an algorithm, thereby providing GTD($\lambda$) algorithm. However, the presented algorithm only covered linear function approximation case and non-linear case was left saying its working is straight forward. However derivation of GTD($\lambda$) even for linear case involved utilizing some properties of linear functions, hence we find it appealing to work with non-linear case as to find how similar properties could be brought up in the case and extend the provided GTD($\lambda$) to non-linear function approximation explicitly.
}

\begin{document}

\maketitle  

\begin{abstract}
Temporal difference (TD) learning is central to reinforcement learning. Many methods use semi-gradient TD, which may diverge, or develop algorithms with mean-squared Bellman or TD-error objectives that are difficult to analyze. Gradient TD algorithms overcome these challenges by using projection of the value function and present convergent off-policy algorithms. However, the algorithm for $\lambda$-return was limited to linear function approximation. Here, we extend the same to non-linear function approximation, which is desirable since many strong modern function approximators are complex and non-linear. Recent works have derived similar algorithms with a generalized projected Bellman error objective. Alternatively, we continue working with the projected Bellman error.
\end{abstract}

\section{Introduction}

Modern reinforcement learning (RL) success in high-dimensional, complex real-world tasks such as robotics and preference learning \citep{intelligence2025, shao2024} is well received. These algorithms use complex non-linear function approximations, making them difficult to analyze and, hence, less well understood. 
Other lines of work present well understood algorithms but are limited in their applicability to high-dimensional real-world applications. Most of these algorithms are restricted to linear function approximation \citep{Maei2011, Antos2007, lagoudakis-2003}.

\cite{sutton-1988} first formalized temporal difference (TD) learning and presented a semi-gradient algorithm. It was also established that for sequential problems with feedback only at the end of the sequence, TD algorithms are more efficient at learning predictions for novel states than supervised learning, provided that predictions for known states are good. This makes TD learning appealing for sequential decision-making, especially since recent advances in pretraining allow reasonable predictions for known states. However, semi-gradient TD algorithms are known to be susceptible to instability and divergence \citep{baird-1995, tsitsiklis-1997}. To address these issues, \cite{Maei2011} introduced gradient TD algorithms that are stable and can learn off-policy. They use a mean-squared projected Bellman error (\textit{MSPBE}) objective to derive the algorithms, which restricts the function class and simplifies analysis. However, the presented gradient TD($\lambda$) algorithm was limited to linear function approximation. More recent works have introduced a generalized variant of the projected Bellman error ($\overline{\text{PBE}}$) that uses a conjugate form of \textit{MSPBE} \citep{Patterson2022}. For convergence analysis, they assume that the value function and the projection space are convex. \cite{Elelimy2025} extend this objective for $\lambda$-return and present a deep reinforcement learning algorithm with gradient eligibility traces. 

As an alternative, we work with \textit{MSPBE} and extend the work of \cite{Maei2011} to non-linear function approximation, presenting an off-policy gradient TD($\lambda$) algorithm in a backward-view eligibility trace formulation with gradient correction and a second-order trace. The derivation assumes that the non-linear value function is locally linear, which makes computing the projection easier. Relaxing this assumption may result in an unanalyzable representation of the projection. We leave convergence and related analysis as a future work.


\section{Definitions}

\subsection{Setup}

We consider a standard RL framework with a stationary MDP $(\mathcal{S, A}, P, R, \gamma)$ and state distribution $\mu(s)$. We consider off-policy data in the form of independent, identically distributed samples $(s,a,r,s')$. The state and action sets are countable (finite or infinite). The goal of an RL agent is to maximize the cumulative reward it receives, which is formulated as a value function. The state value function for a policy $\pi$ is given as 
\[V^\pi(s) = \mathbb{E}\left[\sum_{t=0}^{\infty}\gamma^t R_{t+1} | S_0 = s, \pi\right]\]
Let $P^\pi$ be the state-state transition probability matrix, and $V^\pi$ be a value function vector. $V^\pi$ satisfies the bellman equation $V^\pi = \text{T}^\pi V^\pi = R^\pi + \gamma P^\pi V^\pi$, where T is a bellman operator. RL concerns prediction: evaluating the behavior of a policy, and control: finding an optimal policy. TD is associated with policy evaluation. Hence, we restrict our focus to prediction.

Furthermore, a value function is defined as a non-linear function of the parameters $\theta$ and is denoted by $V_\theta(s_t)$ or $V_t$ for brevity. The TD-error is presented by $\delta_t$ and can be written as $\delta_t = R_{t+1} + \gamma V_{t+1} - V_t$. The corresponding future return is given as $G_t = R_{t+1} + \gamma G_{t+1}$. 
$\lambda$-return formulation allows TD algorithm, effectively TD($\lambda$), a smooth interpolation between single-step backup, TD(0), and monte carlo updates, TD(1). The $\lambda$ weighted TD-error and return take forms, $\delta_t^\lambda = G_t^\lambda - V_t$, and $G_t^\lambda = R_{t+1} + \gamma ((1-\lambda) V_{t+1} + \lambda G_{t+1}^{\lambda})$.

With the off-policy formulation, these can be further defined as $\delta_t^{\lambda\rho} = G_t^{\lambda\rho} - V_t$, and $G_t^{\lambda\rho} = \rho_t (R_{t+1} + \gamma ((1-\lambda) V_{t+1} + \lambda G_{t+1}^{\lambda\rho}))$. Where  $\rho_t =\frac{\pi(a_t|s_t)}{\pi_b(a_t|s_t)}$ is the importance sampling ratio, which accounts for the evaluation of policy $\pi$ with data from policy $\pi_b$.

For tabular RL, $V\in \mathbb{R}^{|\mathcal{S}|}$, i.e., the value function covers all of the real space. However, the purpose of function approximation is to compress the representation, which is critical for generalization and discrimination. This may inadvertently result in a subspace that does not fully cover the required space, where contraction of the Bellman operator can no longer be guaranteed. Using projections and backprojecting values onto the operator space allows for analyzing behavior and deriving guarantees. Alternatives could also include using specific function classes instead of projections.







\subsection{Objective Function}








For any policy $\pi$, the value function satisfies the Bellman equation, such that $V^\pi = \text{T}^\pi V^\pi$. Usually, this relation is used to define an objective function for prediction. The mean squared Bellman error objective does exactly this: $MSBE = ||V - TV||^2_\mu$. However, with function parameterization, it may not be clear how to use this relation, since the fixed point may no longer exist in the parametrized space. Another choice could be to minimize the TD-error appearing in the TD update, resulting in the mean-squared TD-error objective, $MSTDE = \mathbb{E}[\delta_t^2]$. 
More detail on the objective functions can be found in \cite{Maei2011}, who eventually work with 
$MSPBE = ||V_\theta - \Pi \text{T}^{\pi\lambda}V_\theta||^2_\mu$ for TD($\lambda$). Where $\Pi$ is a projection operator used to restrict the resultant transition value in the value function space.

\cite{dai2016} utilize Fenchel Duality and the convex conjugate of $MSBE$ to derive a new objective, $\min_V \max_u 2 (V - TV) u - u^2$, where $u$ is an estimate of $ V$. This idea is formalized for single-step TD learning by \cite{Patterson2022} into a generalized $\overline{\text{PBE}}$ objective. \cite{Elelimy2025} further extend the objective for multi-step TD learning: generalized $\overline{\text{PBE}} = 2\mathbb{E}[\delta^\lambda u] - \mathbb{E}[u^2]$.

The idea of using projection of the value function was first introduced to simplify analysis by restricting updates to a function class \citep{tsitsiklis-1997}. The projected Bellman error objectives have since been well explored for linear function approximation \citep{Antos2007}. 
\cite{Maei2011} use the $MSPBE$ objective to derive off-policy gradient TD algorithms with eligibility trace for linear function approximation. We continue with the same objective, 
\[J(\theta) = || V_{\theta} - \Pi \text{T}^{\pi \lambda}V_{\theta}||^2_{\mu}\]

\section{Non-linear Gradient TD($\lambda$)}

We continue deriving the $\lambda$-return variant of gradient TD algorithms. For brevity, let T = T$^{\pi\lambda}$ and $\nabla =\nabla_\theta$. $D$ is a $\mu$ diagonal matrix. Assume $V_\theta$ is locally linear, the projection $\Pi$ can be written as,
\[\Pi = \nabla V (\nabla V^T D \nabla V )^{-1} \nabla V^T D\]
\[\Pi^T D \Pi = D^T \nabla V (\nabla V^T D \nabla V)^{-1} \nabla V^T D\]
\begin{align*}
    J(\theta) &= (V - \text{T} V)^T D^T \nabla V (\nabla V^T D \nabla V)^{-1} \nabla V^T D (V - \text{T} V) \\
    &= (\nabla V^T D (V - \text{T} V))^T (\nabla V^T D \nabla V)^{-1} \nabla V^T D (V - \text{T} V) \\
    &= \mathbb{E}[\delta^\lambda \nabla V]^T \mathbb{E}[\nabla V \nabla V^T]^{-1} \mathbb{E}[\delta^\lambda \nabla V]
\end{align*}
This is an on-policy formulation, but the data is sampled from $\pi_b$. Hence, we would like to extend this to an off-policy setting. Also note that the middle term is independent of the action.

\textbf{Theorem}: \textit{(Off-policy TD with eligibility traces)} Let $\pi_b$ be a behavior policy and $\delta^\lambda, \delta^{\lambda\rho}$ are as defined above. Then,
\[ \mathbb{E}[\delta^\lambda \nabla V] = \mathbb{E}_{\pi_b}[\delta^{\lambda \rho} \nabla V] \]
\textit{Proof}: See Appendix \ref{off-p-lambda}

Putting all above together, the forward view objective for off-policy learning becomes
\[J(\theta) = \mathbb{E}_{\pi_b}[\delta^{\lambda \rho} \nabla V]^T  \mathbb{E}[\nabla V \nabla V^T]^{-1}  \mathbb{E}_{\pi_b}[\delta^{\lambda \rho} \nabla V] \]
This still requires future rewards to be known, which is easy for a trajectory-based method, instead, we want to derive a single-step algorithm.

\textbf{Theorem}: \textit{(TD forward-view and backward-view objective equivalence)} The forward-view term of the TD update is equivalent to the following backward-view term,
\[\mathbb{E}_{\pi_b}[\delta^{\lambda \rho} \nabla V] = \mathbb{E}_{\pi_b}[\delta e]\]
where $\delta^{\lambda\rho}, \delta$ are as defined above, and $e_t = \rho_t (\nabla V + \gamma \lambda e_{t-1})$

\textit{Proof}: Consider $\delta_t^{\lambda \rho} = G_t^{\lambda \rho} - V_t$, where,
\begin{align*}
    G_t^{\lambda \rho} &= \rho_t (R_{t+1} + \gamma ((1-\lambda) V_{t+1} + \lambda G_{t+1})) \\
    &= \rho_t (R_{t+1} + \gamma ( (1-\lambda) V_{t+1} + \lambda G_{t+1}^{\lambda \rho} \\
    &= \rho_t (R_{t+1} + \gamma V_{t+1} - V_t + V_t) - \rho_t \gamma \lambda V_{t+1} + \rho_t \gamma \lambda G_{t+1}^{\lambda \rho} \\
    G_t^{\lambda \rho} &= \rho_t \delta_t + \rho_t V_t + \rho_t \gamma \lambda \delta_{t+1}^{\lambda \rho}
\end{align*}
\begin{align*}
    \delta_t^{\lambda \rho} &= \rho_t \delta_t + \rho_t V_t + \rho_t \gamma \lambda \delta_{t+1}^{\lambda \rho} \\
    &= \rho_t \delta_t + (\rho_t -1) V_t + \rho_t \gamma \lambda \delta_{t+1}^{\lambda \rho}
\end{align*}
putting above together, 
\[\mathbb{E}[\delta_t^{\lambda \rho} \nabla V_t] = \mathbb{E}[(\rho_t \delta_t + (\rho_t -1) V_t + \rho_t \gamma \lambda \delta_{t+1}^{\lambda \rho})\nabla V_t]\]
The middle term can be independently presented as
\begin{align*}
    \mathbb{E}[(\rho_t-1) V_t \nabla V_t] &= \sum_{s,a} \mu(s) \pi_b(a|s) \left(\frac{\pi(a|s)}{\pi_b(a|s)} - 1\right) V_t \nabla V_t \\
    &= \sum_{s} \mu(s) (\sum_a \pi(a|s) - \sum_a \pi_b(a|s)) V_t \nabla V_t \\
    &= 0
\end{align*}
remember that we were able to take $V_t$ out of the sum over $a$, since it is independent of $a$, this is true for any $V_i$, such that $i \leq t$, which is also referred to as the \textit{causality trick}. 
Also, since the MDP is stationary and the samples are $iid$:
\[\mathbb{E}[\rho_t \gamma \lambda \delta_{t+1}^{\lambda \rho} \nabla V_t] = \mathbb{E}[\rho_{t-1} \gamma \lambda \delta_{t}^{\lambda \rho} \nabla V_{t-1}]\]
\begin{align*}
    \mathbb{E}[\delta_t^{\lambda \rho} \nabla V_t] &= \mathbb{E}[(\rho_t \delta_t +  \rho_t \gamma \lambda \delta_{t+1}^{\lambda \rho})\nabla V_t] \\
    &= \mathbb{E}[\rho_t \delta_t \nabla V_t +  \rho_{t-1} \gamma \lambda \delta_{t}^{\lambda \rho} \nabla V_{t-1}] \\
    &= \mathbb{E}[\rho_t \delta_t \nabla V_t +  \rho_{t-1} \gamma \lambda (\rho_t \delta_t + (\rho_t - 1) V_t + \rho_t \gamma \lambda \delta_{t+1}^{\lambda \rho}) \nabla V_{t-1}] \\
    &= \mathbb{E}[\rho_t \delta_t (\nabla V_t +  \rho_{t-1} \gamma \lambda \nabla V_{t-1}) + 0 + \rho_{t-1} \gamma \lambda \rho_t \gamma \lambda \delta_{t+1}^{\lambda \rho} \nabla V_{t-1}] \\
    &= \mathbb{E}[\rho_t \delta_t (\nabla V_t +  \rho_{t-1} \gamma \lambda \nabla V_{t-1}) + \rho_{t-2} \gamma \lambda \rho_{t-1} \gamma \lambda \delta_{t}^{\lambda \rho} \nabla V_{t-2}] \\
    &= \mathbb{E}[\rho_t \delta_t (\nabla V_t +  \rho_{t-1} \gamma \lambda \nabla V_{t-1}) + \rho_{t-2} \gamma \lambda \rho_{t-1} \gamma \lambda (\rho_t \delta_t + (\rho_t - 1) V_t \\
    &\quad\quad\quad\quad\quad\quad\quad\quad + \rho_t \gamma \lambda \delta_{t+1}^{\lambda \rho}) \nabla V_{t-2}] \\
    &= \mathbb{E}[\rho_t \delta_t (\nabla V_t +  \rho_{t-1} \gamma \lambda \nabla V_{t-1} + \rho_{t-2} \gamma \lambda \rho_{t-1} \gamma \lambda \nabla V_{t-2}) \\
    &\quad\quad\quad\quad\quad\quad\quad\quad + \rho_{t-2} \gamma \lambda \rho_{t-1} \gamma \lambda \rho_t \gamma \lambda \delta_{t+1}^{\lambda \rho} \nabla V_{t-2}] \\
    &= \mathbb{E}[\rho_t \delta_t (\nabla V_t +  \rho_{t-1} \gamma \lambda \nabla V_{t-1} + \rho_{t-1} \gamma \lambda \rho_{t-2} \gamma \lambda \nabla V_{t-2} \\
    &\quad\quad\quad\quad\quad\quad\quad\quad + \rho_{t-1} \gamma \lambda \rho_{t-2} \gamma \lambda \rho_{t-3} \gamma \lambda \nabla V_{t-3} + ... )] \\
    \mathbb{E}[\delta_t^{\lambda \rho} \nabla V_t] &= \mathbb{E}[\delta_t e_t]
\end{align*}
where,
\[e_t = \rho_t (\nabla V_t + \gamma \lambda e_{t-1}) \]
Notice that this new term $\mathbb{E}[\delta_t e_t]$ does not require future returns. 
The objective can equivalently be written as, 
\[J(\theta) = \mathbb{E}[\delta_t e_t]^T \mathbb{E}[\nabla V_t \nabla V_t^T]^{-1} \mathbb{E}[\delta_t e_t]\]
This is the \textit{backward-view objective}.
As our goal is to minimize this objective, we would derive an algorithm to perform an update in the direction of the gradient of $J(\theta)$.
\[\nabla_{\theta} J(\theta) = \nabla_{\theta} \mathbb{E}[\delta_t e_t]^T \mathbb{E}[\nabla V_t \nabla V_t^T]^{-1} \mathbb{E}[\delta_t e_t] \]
assume $A = \mathbb{E}[\delta_t e_t]$, $B = \mathbb{E}[\nabla V_t \nabla V_t^T]$ and $w = B^{-1}A$ 
\begin{align*}
    \nabla J &= \nabla(A^T B^{-1} A) \\
    &= 2 \nabla A^T B^{-1} A + A^T \nabla B^{-1} A \\
    &= 2 \nabla A^T B^{-1} A - A^T B^{-1} \nabla B B^{-1} A \\
    &= 2 \nabla \mathbb{E}[\delta_t e_t]^T w - w^T \mathbb{E}[\nabla(\nabla V_t \nabla V_t^T)] w\\
    &= 2 \nabla \mathbb{E}[\delta_t e_t]^T w - \mathbb{E}[\nabla((w^T \nabla V_t) (\nabla V_t^T w))] \\
    &= 2 \nabla \mathbb{E}[\delta_t e_t]^T w - 2 \mathbb{E}[(\nabla V_t^T w) (\nabla^2 V_t^T w))] \\
    &= \mathbb{E}[\nabla\delta_t e_t^T w] + \mathbb{E}[\delta_t \nabla e_t^T w] - \mathbb{E}[(\nabla V_t^T w) (\nabla^2 V_t^T w))]\\
    &= \mathbb{E}[(\gamma \nabla V_{t+1} - \nabla V_t) e_t^T w] + \mathbb{E}[\delta_t \rho (\nabla^2 V_t^T + \gamma \lambda \nabla e_{t-1}^T)w] - h \\
\end{align*}
where $h$ is an estimate of $\mathbb{E}[(\nabla V_t^T w) (\nabla^2 V_t^T w))]$. Now, consider the term $\nabla e_t^T w$,
\[
    \nabla e_t^T w = \rho (\nabla^2 V_t^T w + \gamma \lambda \nabla e_{t-1}^T w)
\]
let $m_t= \nabla e_t^T w$, then
\[m_t = \rho ( \nabla^2 V_t^T w + \gamma \lambda m_{t-1})\]
putting back into $\nabla J$,
\[\nabla J = \mathbb{E}[(\gamma \nabla V_{t+1} - \nabla V_t) e_t^T w] + \mathbb{E}[\delta_t m_t] - h \]
update for the parameters of $V$ becomes
\[\theta_{t+1} = \theta_t - \alpha [(\gamma \nabla V_{t+1} - \nabla V_t) e_t^T w + \delta_t m_t - h] \]
Recall that $w = \mathbb{E}[\nabla V_t \nabla V_t^T]^{-1} \mathbb{E}[\delta_t e_t]$, which is similar to the normal equation obtained in the regression method. Using
approximate stochastic optimization,
\[w_{t+1} = w_t - \beta (\nabla V_t \nabla V_t^T w_t - \delta_t e_t)\]
Using the equations derived above, a potential algorithm can be written as Alg \ref{alg:gtdnl}.
\begin{algorithm}
\caption{GTD($\lambda$) for Non-linear function approximation}
\label{alg:gtdnl}

\textbf{Initialize} $w_0$ to 0, and $\theta_0$ or $V$ arbitrarily\;
\textbf{Choose} (small) positive values for $\alpha$, $\beta$,
and set values for $\gamma \in (0,1]$, $\lambda \in [0,1]$\;

\textbf{Repeat for each episode:}\;
    \Indp
    \textbf{Initialize} $e_0 = 0, m_0 =0$\;
    \textbf{Take} $A_t$ from $S_t$ according to $\pi_b$, observe $R_{t+1}$ and arrive at $S_{t+1}$\;
    
    \ForEach{step}{
        $\delta_t \gets R_{t+1}
        + \gamma V_{t+1}
        - V_t$\;
        $\rho_t \gets \frac{\pi(A_t|S_t)}{\pi_b(A_t|S_t)}$\;
        $h_t \gets (\nabla V_t^T w_t) (\nabla^2 V_t^T w_t)$\;
        $e_t \gets \rho_t (\nabla V_t + \gamma \lambda e_{t-1})$\;
        $m_t \gets \rho_t (\nabla^2 V_t^T w + \gamma \lambda m_{t-1}$\;
    
        $\theta_{t+1} \gets \theta_t - \alpha [(\gamma \nabla V_{t+1} - \nabla V_t) e_t^T w + \delta_t m_t - h]$\;
        $w_{t+1} \gets w_t - \beta (\nabla V_t \nabla V_t^T w_t - \delta_t e_t)$\;
    }
    \Indm
\end{algorithm}

During implementation of the algorithm, computational improvements could be possible. e.g., for efficiency, in $m_t$, computing $\nabla^2 V_t^T w$ as $\nabla ( \nabla V_t^T w)$ removes the need for matrix-vector product. This is possible since $w$ is an estimate and independent of the parameters.

\section{Discussion}

We derived a gradient TD algorithm with $ \ lambda$-return for non-linear function approximation under the local linearity assumption using the MSPBE objective. This paves the way for (i) implementing GTD($\lambda$) with non-linear functions, specifically useful for implementing off-policy actor-critic methods with more capable functions, (ii) analyzing the behavior of the algorithm and checking its convergence. Further, other interesting directions include enforcing local linearity assumptions on the function class using methods such as Lipschitz continuity and smaller parameter updates. It can be considered that enforcing convexity may be simpler, in which case methods that work with the generalized $\overline{\text{PBE}}$ objective could be useful.

\bibliography{main}
\bibliographystyle{rlj}

\appendix
\section*{Appendix}
This section elaborates some of the assumptions and definitions used in the main text.

\section{Deriving Projections}
\label{projections}

Assume the value function is locally linear. We can find the projection of T$V_{\theta}$ on $V_{\theta}$. Let $v = \text{T}V_\theta$, using
\[\Pi v = V_{\theta^\star}\] 
locally, $V_\theta$ can be represented directly in the form of its parameters as
\[V_\theta = \nabla V_\theta \theta\]
where, 
\[\theta^\star = \arg \min_\theta ||V_\theta - v||^2_\mu\]
\[\theta^\star = \arg \min_\theta (V_\theta - v)^T D (V_\theta - v)\]
Differentiate to find $\theta^\star$
\[2 \nabla V_\theta^T D (V_{\theta^\star} -v) = 0\]
\[\nabla V_\theta^T D \nabla V^T \theta^\star = \nabla V_\theta^T D v\]
\[\theta^\star = (\nabla V^T D \nabla V)^{-1} \nabla V^T D v\]
Putting back in $\Pi v$ to back-project the error on $V$, we get a closed-form formulation of the projection
\[ \Pi = \nabla V (\nabla V^T D \nabla V)^{-1} \nabla V^T D \]
Using this, $\Pi^T D \Pi $ could be further written as,
\begin{align*}
    \Pi^T D \Pi &= (\nabla V (\nabla V^T D \nabla V)^{-1} \nabla V^T D)^T D \nabla V (\nabla V^T D \nabla V)^{-1} \nabla V^T D \\
    &= D^T \nabla V (\nabla V^T D \nabla V)^{-1} \nabla V^T D \nabla V (\nabla V^T D \nabla V)^{-1} \nabla V^T D \\
    &= D^T \nabla V (\nabla V^T D \nabla V)^{-1} \nabla V^T D
\end{align*}
Note that $\nabla V^T D \nabla V$ is symmetric

\section{Off-policy TD with eligibility traces}
\label{off-p-lambda}
\textbf{Theorem}: Let $\pi_b$ be a behavior policy and $\delta^\lambda, \delta^{\lambda\rho}$ are as defined above, then,
\[ \mathbb{E}[\delta^\lambda \nabla V] = \mathbb{E}_{\pi_b}[\delta^{\lambda \rho} \nabla V] \]
\textit{Proof}:
We derive the equivalence by solving the $RHS$
\[\mathbb{E}_{\pi_b}[\delta^{\lambda\rho} \nabla V_t] = \mathbb{E}_{\pi_b}[(G_t^{\lambda\rho} - V_t) \nabla V_t]\]
Notice, $V_t$ is independent of $\pi_b$, solving for $G_t^{\lambda \rho}$
\begin{align*}
    \mathbb{E}_{\pi_b}[G_t^{\lambda\rho}] &= \mathbb{E}_{\pi_b}[\rho_t(R_{t+1} + \gamma (1-\lambda) V_{t+1}) + \rho_t \gamma \lambda G_{t+1}^{\lambda\rho} | S_t = s] \\
    &= \mathbb{E}_{\pi_b}[\rho_t(R_{t+1} + \gamma (1-\lambda) V_{t+1})] + \rho_t \gamma \lambda \mathbb{E}[G_{t+1}^{\lambda\rho} | S_t = s] \\
    &= \mathbb{E}[R_{t+1} + \gamma (1-\lambda) V_{t+1} | S_t = s, \pi] + \sum_{a,s'} p(s'|s,a) \pi_b(a|s) \rho_t \gamma \lambda \mathbb{E}[G_{t+1}^{\lambda\rho} | S_{t+1} = s'] \\
    &= \mathbb{E}[R_{t+1} + \gamma (1-\lambda) V_{t+1} | S_t = s, \pi] + \sum_{a,s'} p(s'|s,a) \pi(a|s) \gamma \lambda \mathbb{E}[G_{t+1}^{\lambda\rho} | S_{t+1} = s'] \\
    \mathbb{E}_{\pi_b}[G_t^{\lambda\rho}] &= \mathbb{E}[R_{t+1} + \gamma (1-\lambda) V_{t+1} + \gamma \lambda \mathbb{E}[G_{t+1}^{\lambda\rho} | S_{t+1} = s'] | S_t = s, \pi] \\
    &= \mathbb{E}[R_{t+1} + \gamma (1-\lambda) V_{t+1} + \gamma \lambda \mathbb{E}[R_{t+2} + \gamma (1-\lambda) V_{t+2} \\
    &\quad\quad\quad\quad\quad\quad\quad\quad + \gamma\lambda \mathbb{E}[G_{t+2}^{\lambda\rho} | S_{t+2} = s'']|S_{t+1}=s', \pi] | S_t = s, \pi] \\
    &= \mathbb{E}[R_{t+1} + \gamma (1-\lambda) V_{t+1} + \gamma \lambda (R_{t+2} + \gamma (1-\lambda) V_{t+2} \\
    &\quad\quad\quad\quad\quad\quad\quad\quad  + \gamma\lambda \mathbb{E}[G_{t+2}^{\lambda\rho} | S_{t+2} = s'']) | S_t = s, \pi] \\
    &... \\
    &= \mathbb{E}[G_t^\lambda | S_t = s, \pi]
\end{align*}
Putting back in $RHS$, 
\begin{align*}
    \mathbb{E}_{\pi_b}[\delta^{\lambda\rho} \nabla V_t] &= \mathbb{E}_{\pi_b}[(G_t^{\lambda\rho} - V_t) \nabla V_t] \\
    &= \mathbb{E}[(G_t^\lambda -V_t) \nabla V_t | S_t=s,\pi] \\
    &= \mathbb{E}_\pi[\delta^\lambda_t\nabla V_t]\\
\end{align*}

\end{document}